\documentclass[conference]{IEEEtran}
\IEEEoverridecommandlockouts
\usepackage{cite}
\usepackage{amsmath,amssymb,amsfonts}
\usepackage{algorithmic}
\usepackage{graphicx}
\usepackage{textcomp}
\usepackage{xcolor}
\usepackage{bm}
\usepackage{multirow}
\usepackage{balance}
\usepackage{colortbl}
 \usepackage{float}
\IEEEoverridecommandlockouts                           
\makeatletter
\let\NAT@parse\undefined
\makeatother
\usepackage{hyperref}
\hypersetup{colorlinks=true, linkcolor=red, citecolor=blue, urlcolor=magenta}

\def\BibTeX{{\rm B\kern-.05em{\sc i\kern-.025em b}\kern-.08em
    T\kern-.1667em\lower.7ex\hbox{E}\kern-.125emX}}
\begin{document}

\title{\LARGE \bf
Towards Precise 3D Human Pose Estimation with Multi-Perspective Spatial-Temporal Relational Transformers
}

\author{
    \IEEEauthorblockN{
        Jianbin Jiao\textsuperscript{1}, 
        Xina Cheng\textsuperscript{1}$^{,*}$, 
        Weijie Chen\textsuperscript{1}, 
        Xiaoting Yin\textsuperscript{2}, 
        Hao Shi\textsuperscript{2}, and Kailun Yang\textsuperscript{3,4} 
    }
    \IEEEauthorblockA{\textsuperscript{1}School of Artificial Intelligence, Xidian University, China\\}
    \IEEEauthorblockA{\textsuperscript{2}State Key Laboratory of Extreme Photonics and Instrumentation, Zhejiang University, China\\}
    \IEEEauthorblockA{\textsuperscript{3}School of Robotics, Hunan University, China.\\}
    \IEEEauthorblockA{\textsuperscript{4}National Engineering Research Center of Robot Visual Perception and Control Technology, Hunan University, China.\\}
    Correspondence: xncheng@xidian.edu.cn
}

\maketitle
\thispagestyle{empty}
\pagestyle{empty}

\begin{abstract}
3D human pose estimation captures the human joint points in three-dimensional space while keeping the depth information and physical structure. That is essential for applications that require precise pose information, such as human-computer interaction, scene understanding, and rehabilitation training. Due to the challenges in data collection, mainstream datasets of 3D human pose estimation are primarily composed of multi-view video data collected in laboratory environments, which contains rich spatial-temporal correlation information besides the image frame content. Given the remarkable self-attention mechanism of transformers, capable of capturing the spatial-temporal correlation from multi-view video datasets, we propose a multi-stage framework for 3D sequence-to-sequence (seq2seq) human pose detection. Firstly, the spatial module represents the human pose feature by intra-image content, while the frame-image relation module extracts temporal relationships and 3D spatial positional relationship features between the multi-perspective images. Secondly, the self-attention mechanism is adopted to eliminate the interference from non-human body parts and reduce computing resources. Our method is evaluated on Human3.6M, a popular 3D human pose detection dataset. Experimental results demonstrate that our approach achieves state-of-the-art performance on this dataset. The source code will be available at \url{https://github.com/WUJINHUAN/3D-human-pose}.

\end{abstract}


\section{Introduction}

Human pose estimation provides the positions of human key joint points from images or videos, which is the core technology for understanding human behaviors and motion scenes. 
The reliability and generalization of these applications are highly reliant on the precision of human pose data, especially in human-computer interaction, virtual reality, and rehabilitation training, the 3D human pose is required. 
The primary task of 3D human pose estimation is to predict the three-dimensional coordinates of human body keypoints in the 3D space.
Compared with the image-based 2D pose estimation, which has been quite matured thanks to the powerful deep learning method~\cite{wei2016convolutional,xu2022vitpose,cao2017realtime,sun2019deep}, 3D human pose estimation~\cite{pavlakos2017coarse,chen2022efficient,qiu2019cross,martinez2017simple} provides 3D keypoints positions. With the depth information, the 3D human pose is represented keeping the physical structure and connected relation. The rich data is essential for applications requiring precise pose information.  

This paper targets the 3D human pose estimation from the multi-view video data, which is the mainstream dataset of 3D pose~\cite{mehta2017monocular,ionescu2013human3,sigal2010humaneva}. 
Besides the image frame content, these datasets provide temporal information and 3D spatial relations. 
Numerous studies aim to leverage spatial-temporal information from these datasets to achieve enhanced performance of human pose detection~\cite{liu2020attention,chen2021anatomy,hossain2018exploiting}, which are broadly categorized into two types according to the backbone network: the CNN based method ~\cite{pavllo20193d} and Transformer based method~\cite{li2021tokenpose,yang2021transpose,li2021pose}. 
Given the richness of multi-view information and temporal relation in 3D human pose datasets, the transformer shows the strength to model long-range dependencies in data with attention mechanisms.
However, current transformer-based methods primarily investigate image features without harnessing the power of transformers to extract rich feature information from video datasets~\cite{mehta2017monocular,ionescu2013human3,sigal2010humaneva}. For instance, PRTR~\cite{li2021pose} adopts a structure comprising both transformer encoders and decoders, cascadingly refining the estimation of keypoint positions. Conversely, TokenPose~\cite{li2021tokenpose} and TransPose~\cite{yang2021transpose} utilize a transformer architecture with only encoders to process features extracted by CNNs. These approaches fail to effectively exploit the rich feature information in the dataset and overcome the limitations of CNN-based methods, which solely focus on intrinsic image features.

\begin{figure}[t]
  \centering
  \includegraphics[width=0.5\textwidth]{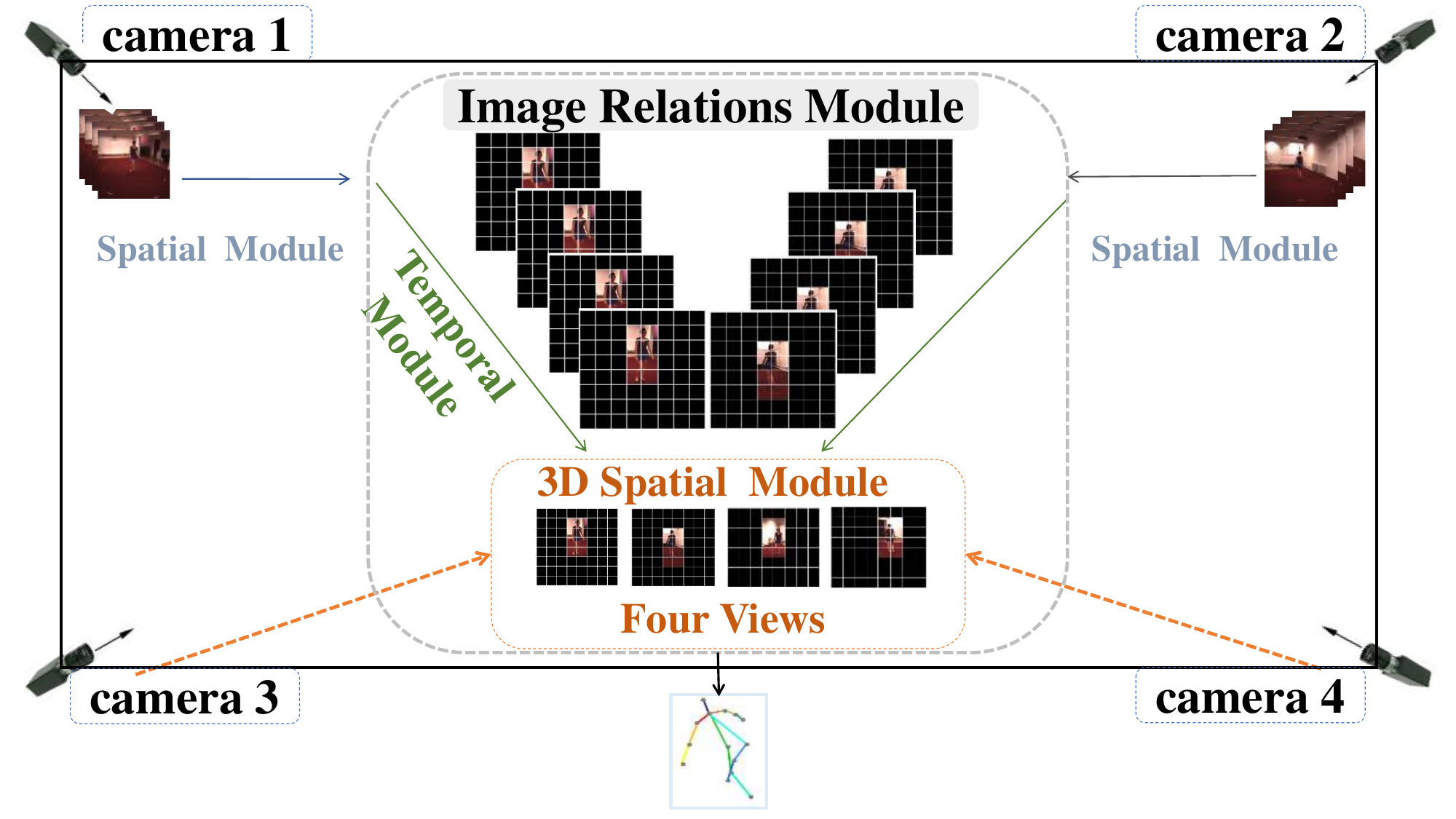} %
  \caption{The conceptual diagram of our approach consists of two main components: the Spatial Module and the Image Relations Module. The Spatial Module extracts human body pose features inherent in the images themselves, while the Image Relations Module first models the temporal relationships between frame images and subsequently models the spatial positional relationships between corresponding images in 3D space.} %
  \label{img1} %
  \vspace{-0.4cm}
\end{figure}

Based on the identified shortcomings and deficiencies mentioned above, our network architecture is founded on the self-attention mechanism of the transformer, enabling the comprehensive extraction of various rich feature information in 3D human pose estimation. Concerning input data, our network takes all frames of a video sequence and feeds them into the transformer architecture, facilitating the direct construction of spatial and temporal information for video frames. Our network comprises two components: the spatial module and the frame-image relation module.

Fig.~\ref{img1} illustrates the structure of our proposed approach. 
The effectiveness of this approach is evaluated on the widely adopted dataset for 3D human body pose estimation: Human3.6M~\cite{ionescu2013human3}.
As the Human3.6M dataset~\cite{ionescu2013human3} comprises videos captured from four cameras, it inherently contains image information, 3D spatial positional information, and temporal information.
The spatial module is designed to extract the image feature and the images relation module is for harvesting the 3D spatial and temporal features.

Firstly, the spatial module is responsible for gathering internal information within video frames.
This section involves extracting feature information from images to construct intra-frame spatial relationships between keypoints. 
Considering that the Human3.6M~\cite{ionescu2013human3} dataset consists of data extracted from the same scene, we perform image cropping in this phase.
A windowed self-attention mechanism is employed as the backbone to retain image patches with high attention and eliminate the disturbance of non-human body patches.
Simultaneously, this approach has the ability to reduce the computational load caused by long sequences of input images. 

Secondly, the images relation module involves modeling the relationships between frame images.
On the time scale, the output image features of the spatial module are treated as a token input to model the temporal relationships within a sequence of frame images.
On a 3D spatial scale, the 3D spatial positional relationships are modeled between images.
In this part, we model the 3D spatial positional information between images. Given that the 3D spatial and temporal features require modeling global relationships, we employ a basic transformer model for this purpose. Our contributions are summarized into three key aspects:
\begin{itemize}
\item We introduce a novel 3D sequence-to-sequence human pose detection network that incorporates intra-frame spatial, temporal, and 3D position information. Notably, we applied window self-attention for the first time in frame-based human pose detection. Moreover, we combined window self-attention with global self-attention, effectively reducing computational complexity while modeling global relationships. This integration maximizes the advantages of both structures.
\item In the spatial domain, we implement the cropping of image patches, selectively preserving patches relevant to pose detection. This approach led to an improvement in network performance by focusing on pertinent information within the image.
\item In the temporal domain, we propose to model the temporal information between video frames and extract corresponding 3D spatial position information. This comprehensive consideration of temporal dynamics and spatial relationships between frames contributes to a more robust understanding of the context in video-based human pose detection.
\end{itemize}

\section{Related Work}
In this part, we present a review of relevant literature on 2D and 3D human pose estimation, discussing the insights derived from these methods, and elucidating the distinctions from our proposed approach. 
Subsequently, we delve into the impact of existing research on vision transformers, highlighting how these studies have influenced our methodology.

\subsection{2D Human Pose Detection}
Before the widespread application of transformers in various image tasks, most 2D human keypoint detection methods relied on Convolutional Neural Networks (CNNs) as the main backbone.
These 2D human keypoint detection approaches are categorized into multi-person and single-person detection based on image content. 
Multi-person human keypoint detection is further divided into top-down~\cite{chen2018cascaded,newell2016stacked,xiao2018simple,li2019rethinking} and bottom-up~\cite{cao2017realtime,sun2019deep} approaches. 
In the bottom-up approach, all keypoints of all individuals are detected first, followed by using clustering algorithms to classify keypoints belonging to the same person, resulting in the final output. 
On the other hand, the top-down approach involves using object detection algorithms~\cite{carion2020end,he2017mask} to obtain bounding boxes for each person and then performing single-person keypoint detection based on these bounding boxes. 
Among these methods, the top-down approach consistently achieves state-of-the-art results on benchmark datasets, leading to the choice of employing the top-down approach in this work.

While CNNs and their variants remain the primary backbone architecture for computer vision applications, the modeling capabilities of transformers~\cite{vaswani2017attention,dosovitskiy2020image} for various information relationships are unparalleled by CNNs. 
In this context, we aim to leverage the transformer architecture to extract rich feature information from 3D human pose datasets.
Despite the continued prominence of CNN models, the unique ability of transformers to harvest complex relationships motivates our exploration of their potential in capturing intricate features within 3D human pose data.

\subsection{3D Human Pose Estimation}
3D human pose recognition can be broadly categorized into direct methods~\cite{pavlakos2017coarse} and two-stage methods~\cite{chen2022efficient,qiu2019cross,zheng2021_3d_hpe}.
Direct methods aim to extract raw feature information directly from images. 
For instance, C2F-Vol~\cite{pavlakos2017coarse} draws inspiration from the Hourglass network structure employed in 2D Human Pose Estimation (HPE) and represents 3D pose in the form of 3D heatmaps. 
Two-stage methods, on the other hand, simplify the prediction process.
The baseline approach utilizes a relatively simple feedforward neural network for estimating 3D pose, taking 2D keypoint coordinates as input and directly mapping 2D pose to 3D space through fully connected layers with residual connections. 
Hossain and Little~\cite{hossain2018exploiting} introduce a Recurrent Neural Network (RNN) with Long Short-Term Memory (LSTM) units to leverage temporal information within input sequences.
Pavllo~\textit{et al.}~\cite{pavllo20193d} incorporate a temporal convolutional network to estimate 3D pose generated from continuous 2D sequences of keypoints. 
In~\cite{chen2022efficient,yin2023rethinking}, efficient human pose estimation has been explored by using event point cloud data.
However, these networks~\cite{hossain2018exploiting,pavllo20193d} often struggle to model information relationships between frames in multi-view video sequences effectively. 
To address this limitation, we employ a transformer structure to extract information from multi-view video datasets. 
Leveraging rich feature information, this approach yields accurate 2D human pose estimations. 
Finally, a two-stage method is employed to achieve 3D human pose recognition.

\subsection{Vision Transformer}
After the initial introduction of Transformers for image tasks, numerous variant architectures suitable for various image-related tasks have been proposed. 
DETR~\cite{carion2020end}, applied in object detection, achieved an end-to-end framework for object detection. 
ViT~\cite{dosovitskiy2020image} successfully employed a pure Transformer structure for image classification, demonstrating excellent performance. Given the substantial computational resources required for self-attention calculations on images, several models focusing on reducing computational complexity have been introduced. 
The Swin transformer~\cite{liu2021swin} divides images into small windows, performing self-attention computations within these windows. 
In~\cite{zeng2022not}, the presented TCFormer achieves sparse input by clustering and pruning the input tokens. 
DynamicViT~\cite{rao2021dynamicvit} evaluates tokens generated for an image, discarding low-scoring background tokens to facilitate sparse input. 
Moreover, the transformer architecture has been introduced to various tasks like semantic segmentation~\cite{xie2021segformer,zhang2022trans4trans}, content outpainting~\cite{shi2023fishdreamer,shi2022flowlens} and pose estimation~\cite{xu2022vitpose,zheng2021_3d_hpe,li2022mhformer,li2022exploiting}.
In this work, our proposed network architecture leverages windowed self-attention to efficiently extract spatial information regarding human key points in the image.
Furthermore, we introduce an image relations module to harvest temporal relationship features among frames in video sequences and 3D spatial position features for enhancing the estimation performance.

\begin{figure*}[h]
  \centering
  \includegraphics[width=1\textwidth]{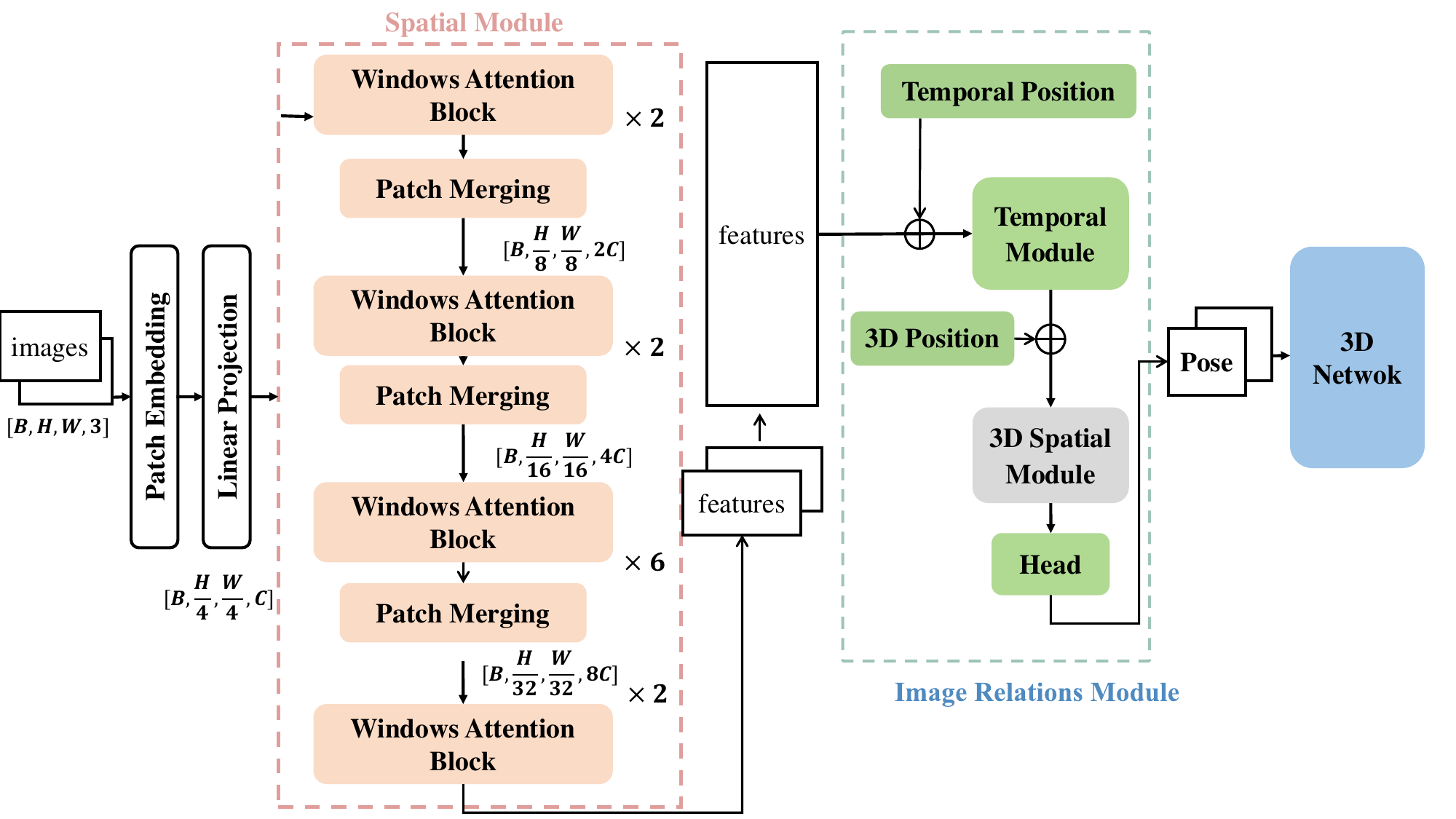} %
  \caption{A network architecture is proposed for extracting 2D human body poses using the self-attention mechanism, comprising two modules. The Spatial Module extracts pose features from images with windowed self-attention. The Image Relations Module extracts temporal relationships and 3D spatial features from video frames, using global self-attention to learn sequence-wide relationships. The final output is the 2D human body pose, utilized to estimate 3D poses.} %
  \label{img2} %
  \vspace{-0.3cm}
\end{figure*}

\section{Method}
\subsection{Overview Architecture}

In this section, we introduce a multi-stage framework for
3D sequence-to-sequence (seq2seq) human pose detection,
as illustrated in Fig.~\ref{img2}. Our architecture primarily comprises
a spatial module and a frame-image relation module, both
adeptly leveraging the self-attention mechanism~\cite{vaswani2017attention}. The spatial module is utilized to extract human body pose features inherent in the images themselves, while the frame-image relation module extracts temporal relationships and 3D spatial positional relationship features between the images. Connecting
these two components is a frame-image information aggregation module, responsible for aggregating all tokenized information containing image details into a sequence of
tokens that represent the frames within a video. Finally, keypoint coordinates for all input frames are directly estimated
through a regression head, in accordance with the standard
seq2seq architecture. 

\subsection{Motivation}

The fundamental premise of our research centers on the self-attention mechanism~\cite{vaswani2017attention}, an overarching architecture adept at modeling dependencies within input sequences. Initially, the input images $I$ undergo processing through embedding~\cite{dosovitskiy2020image} operations,
\begin{equation}
    \begin{aligned}
       Z=Patch\ Embedding(I),
    \end{aligned}
\end{equation}
where $Z\in \mathbb{R} ^{N \times C}$ represents the patch sequence, with $N$ being the length of the patch sequence and $C$ being the dimensionality of each patch.
Subsequently, the self-attention mechanism transforms $Z\in \mathbb{R} ^{N \times C}$ into three vectors, $Q$, $K$, and $V$, through three distinct linear transformations:

\begin{equation}
    \begin{aligned}
       Q = ZW_Q, K = ZW_K, V = ZW_V. 
    \end{aligned}
\end{equation}
Among these, $Q$ can be understood as the information to be queried, $K$ represents the patches to be queried, and $V$ signifies the features of each patch. Following the derivation of these three vectors, the computation of self-attention can be expressed as:
\begin{equation}
    \begin{aligned}
       Attention(Q, K, V) = Softmax(\frac{Q{\cdot}K}{\sqrt{d}})V,
    \end{aligned}
\end{equation}
where$\sqrt{d}$ serves as a normalization factor for the values of $Q{\cdot}K$, preventing numerical instability that may lead to gradient vanishing. The $Softmax(\cdot)$ function normalizes the computed results, generating a weight matrix that, when multiplied by the feature vectors $V$, yields the attention values. To capture intricate feature information, the self-attention mechanism employs a multi-head self-attention layer ($MSA$) for parallel processing of feature information. Each head simultaneously computes self-attention, and the outputs from $k$ heads are concatenated. This can be expressed as:
\begin{equation}
    \begin{aligned}
       MultiHead(Q, K, V) = Concat(head_1, ..., head_h)W_{Out}
    \end{aligned}
\end{equation}
\begin{equation}
    \begin{aligned}
       where\ head_i = Attention(Q_i, K_i, V_i), i\in \mathbb [1, ..., h]
    \end{aligned}
\end{equation}

In 3D pose estimation datasets, rich feature information is present, and we aim to model the specific dependencies among these features. Therefore, we employ the self-attention mechanism as the core mechanism in our network.

\subsection{Spatial Module}
\begin{figure}[h]
  \centering
  \includegraphics[width=0.5\textwidth]{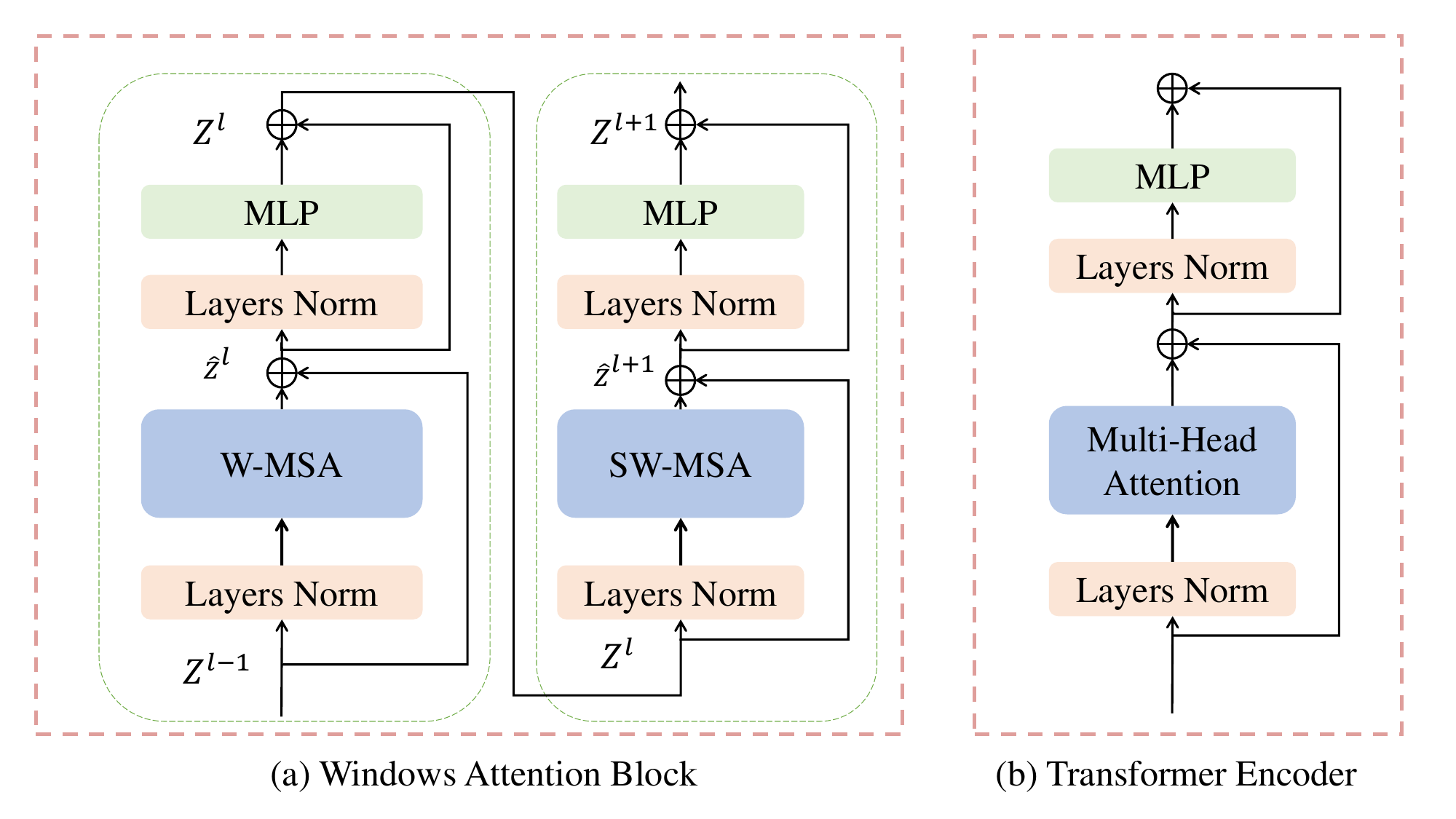} %
  \caption{(a) Mobile Window Attention Module: This module limits self-attention computations to a small window. To capture global features, the window moves across the image, conducting self-attention computations within each image block generated after each movement. (b) Standard Transformer Module.} %
  \label{img3} %
  \vspace{-0.4cm}
\end{figure}

As a network tailored for processing video sequences, we need to design a spatial feature extraction module capable of handling long input video frame sequences.
Given the substantial volume of video frame data, we employ a moving window self-attention mechanism to extract spatial features, which helps alleviate the computational burden and reduce processing time.
Specifically, we partition a single image into multiple small blocks using a small window, limiting self-attention computations to the interior of these blocks. Simultaneously, we incorporate window movement to capture global features, facilitating interaction with image information outside the current window.
Subsequently, within the generated image blocks after movement, another round of self-attention computation is performed. Consequently, as illustrated in Fig.~\ref{img3}(a), a complete spatial transformer block involves two rounds of self-attention computation, which can be expressed as: 
\begin{equation}
    \begin{aligned}
        &\hat{z}^l=W-MSA(LN(z^{l-1})+z^{l-1}
    \end{aligned}
\end{equation}
\begin{equation}
    \begin{aligned}
        &z^l=MLP(LN(\hat{z}^l))+\hat{z}^l
    \end{aligned}
\end{equation}
\begin{equation}
    \begin{aligned}
        &\hat{z}^{l+1}=SW-MSA(LN(z^l)+z^l
    \end{aligned}
\end{equation}
\begin{equation}
    \begin{aligned}
        &z^{l+1}=MLP(LN(z^{l+1})+\hat{z}^{l+1}
    \end{aligned}
\end{equation}

In this context, $\hat{z}^l$ and $z^l$ respectively denote the conventional window module of self-attention blocks and the output features of the $MLP$ module. Here, $W-MSA$ and $SW-MSA$ refer to utilizing conventional and sliding window partitions in window-based self-attention.

To enable the network to learn hierarchical features, a patch merging module performs downsampling on the images. This aids in reducing the length of the patch sequence, increasing the patch dimension, and enlarging the receptive field for extracting contextual cues. 
The downsampling operation generates images at different scales, facilitating the network in learning features from multiple scales.

\subsection{Image Relations Module}

Following the extraction of image spatial features using the spatial module, we consolidate the information of each image into a vector that encapsulates the spatial characteristics of the image.
Subsequently, we concatenate the feature information of each frame's image as tokens and input them into the standard transformer model.

When modeling the relationships between images, as illustrated in Fig.~\ref{img3}(b), we employ a standard transformer model~\cite{vaswani2017attention}, as a significant amount of spatial image feature information has already been consolidated into a single vector during the spatial module, facilitating the flexible computation of global self-attention. 
In a standard transformer model (assuming input of $32$ frames), if the input frame image size is $(32, 224, 224, 3)$, after mapping, the size of tokens input to the transformer module is $(32, 196, 768)$. However, in our structure, after the spatial feature extraction, the size of tokens input to the transformer model is $(1, 32, 768)$. This significantly reduces computational complexity, enabling effortless calculation of global self-attention. The Image Relations Module is divided into two parts. The first part models the temporal relationships between video frames, utilizing self-attention mechanisms to learn global dependencies between video frames. The second part models the 3D spatial positional relationships between corresponding images. Since data collection involves the use of four cameras, a 3D pose corresponds to four 2D poses, and there exists a spatial positional relationship between these four 2D poses. We utilize self-attention mechanisms to learn this relationship. Additionally, sometimes, to fully leverage temporal information in video sequences, long sequences of video frames may be inputted, a scenario our model can readily accommodate.

Our Image Relations Module can be represented as follows:
\begin{equation}
    \begin{aligned}
    {Z}_{l}^{'}=MSA(LN(Z_{l-1}))+Z_{l-1},&l=1,2...L
    \end{aligned}
\end{equation}
\begin{equation}
    \begin{aligned}
    Z_l=MLP(LN({Z}_{l}^{'}))+{Z}_{l}^{'},&l=1,2,...L
    \end{aligned}
\end{equation}
\begin{equation}
    \begin{aligned}
    Y=LN(Z_L)
    \end{aligned}
\end{equation}
where $MSA$ represents the multi-head self-attention module, $LN(\cdot)$~\cite{ba2016layer} represents the normalization layer, $Y$ is the output, and $Z$ represents the output of each layer of the transformer module, with a total of $L$ layers.

We employ positional embeddings~\cite{dosovitskiy2020image} to represent the temporal dependencies and spatial relationships in the 3D space of the images:
\begin{equation}
    \begin{aligned}
    &Z_{time} = [z_1;z_2;. . .;z_f] + E_{time}   
    \end{aligned}
\end{equation}
\begin{equation}
    \begin{aligned}
    &Z_{3D}=[z_1;z_2;z_3;z_4] + E_{3D}
    \end{aligned}
\end{equation}
where $z$ represents the feature information of each image, $E_{time}$ and $E_{3D}$ denote two types of relationships.

\begin{table*}[h]
\scriptsize
\renewcommand{\thetable}{II}
\centering
\caption{Evaluation Results for 3D Human Pose}
\resizebox{\linewidth}{!}{
\setlength{\tabcolsep}{3pt}
\begin{tabular}{llcccccccccccccccc}
\hline
\textbf{MPJPE$\downarrow$}       &   \textbf{Venue}        & \textbf{Dir.} & \textbf{Disc.} & \textbf{Eat}  & \textbf{Greet} & \textbf{Phone} & \textbf{Photo} & \textbf{Pose} & \textbf{Purch.} & \textbf{Sit}  & \textbf{SitD.} & \textbf{Somke} & \textbf{Wait} & \textbf{WalkD.} & \textbf{Walk} & \textbf{WalkT.} & \textbf{Average} \\
\hline
Dabral \textit{et al.}~\cite{dabral2018learning} &  {ECCV’18 } & 44.8 & 50.4  & 44.7 & 49.0  & 52.9  & 61.4  & 43.5 & 45.5   & 63.1 & 87.3  & 51.7  & 48.5 & 52.2   & 37.6 & 41.9   & 52.1    \\
Cai \textit{et al.}~\cite{cai2019exploiting}   &  {ICCV’19}  & 44.6 & 47.4  & 45.6 & 48.8  & 50.8  & 59.0  & 47.2 & 43.9   & 57.9 & 61.9  & 49.7  & 46.6 & 51.3   & 37.1 & 39.4   & 48.8    \\
Pavllo \textit{et al.}~\cite{pavllo20193d} & { CVPR’19}  & 45.2 & 46.7  & 43.3 & 45.6  & 48.1  & 55.1  & 44.6 & 44.3   & 57.3 & 65.8  & 47.1  & 44.0 & 49.0   & 32.8 & 33.9   & 46.8    \\
Lin \textit{et al.}~\cite{lin2019trajectory}  &  {BMVC’19}  & 42.5 & 44.8  & 42.6 & 44.2  & 48.5  & 57.1  & 52.6 & 41.4   & 56.5 & 64.5  & 47.4  & 43.0 & 48.1   & 33.0 & 35.1   & 46.6    \\
Yeh \textit{et al.}~\cite{yeh2019chirality}   &  {NIPS’19}  & 44.8 & 46.1  & 43.3 & 46.4  & 49.0  & 55.2  & 44.6 & 44.0   & 58.3 & 62.7  & 47.1  & 43.9 & 48.6   & 32.7 & 33.3   & 46.7    \\
Liu \textit{et al.}~\cite{liu2020attention} & { CVPR’20}  & 41.8 & 44.8  & 41.1 & 44.9  & 47.4  & 54.1  & 43.4 & 42.2   & 56.2 & 63.6  & 45.3  & 43.5 &{\color{blue}45.3}  &{\color{blue}31.3} & 32.2   & 45.1    \\
SRNet~\cite{zeng2020srnet}      &  {ECCV’20}  & 46.6 & 47.1  & 43.9 &{\color{blue}41.6}  &{\color{blue}45.8}  &{\color{blue}49.6}  & 46.5 &{\color{blue}40.0}  &{\color{blue}53.4} & 61.1  & 46.1  & 42.6 & {\color{blue}43.1} & 31.5 & 32.6   & 44.8    \\
UGCN~\cite{wang2020motion}       &  {ECCV’20}  &{\color{blue}41.3} & 43.9  & 44.0 & 42.2  & 48.0  & 57.1  & 42.2 & 43.2   & 57.3 & 61.3  & 47.0  & 43.5 & 47.0   & 32.6 &{\color{blue}31.8}   & 45.6    \\
Chen \textit{et al.}~\cite{chen2021anatomy}&TCSVT’21 & 42.1 &{\color{blue}43.8} & 41.0 & 43.8  & 46.1  & 53.5  & 42.4 & 43.1   & 53.9 &{\color{blue}60.5} & 45.7  &{\color{blue}42.1} & 46.2   & 32.2 & 33.8   & 44.6    \\
\rowcolor{gray!25}
PoseFormer~\cite{zheng2021_3d_hpe} &ICCV’21   & 41.5 & 44.8  &{\color{blue}39.8} & 42.5  & 46.5  & 51.6  &{\color{blue}42.1} & 42.0   &{\color{red}53.3} & 60.7  & 45.5  & 43.3 & 46.1   & 31.8 & 32.2   &{\color{blue}44.3}   \\
\rowcolor{gray!25}
Ours &          &{\color{red}35.2}&{\color{red}41.0} &{\color{red}37.9} &{\color{red}36.9} & {\color{red}39.4} &{\color{red}45.7}  &{\color{red}38.7} &{\color{red}38.7}   &54.4 &{\color{red}58.2}  &{\color{red}40.8}  &{\color{red}38.8} & {\color{red}41.0}   &{\color{red}27.5} & {\color{red}29.5}   & {\color{red}40.3}    \\

\hline
\hline

\textbf{P-MPJPE$\downarrow$} &   \textbf{Venue}        & \textbf{Dir.} & \textbf{Disc.} & \textbf{Eat}  & \textbf{Greet} & \textbf{Phone} & \textbf{Photo} & \textbf{Pose} & \textbf{Purch.} & \textbf{Sit}  & \textbf{SitD.} & \textbf{Somke} & \textbf{Wait} & \textbf{WalkD.} & \textbf{Walk} & \textbf{WalkT.} & \textbf{Average} \\
\hline
Pavlakos \textit{et al.}~\cite{pavlakos2018ordinal} &  {CVPR’18} & 34.7 & 39.8  & 41.8 & 38.6  & 42.5  & 47.5  & 38.0 & 36.6   & 50.7 & 56.8  & 42.6  & 39.6 & 43.9  & 32.1  & 36.5   & 41.8 \\
Hossain \textit{et al.}~\cite{hossain2018exploiting} &  {ECCV’18} & 35.7 & 39.3  & 44.6 & 43.0  & 47.2  & 54.0  & 38.3 & 37.5   & 51.6 & 61.3  & 46.5  & 41.4 & 47.3  & 34.2  & 39.4   & 44.1 \\
Cai \textit{et al.}~\cite{cai2019exploiting} &  {ICCV’19} & 35.7 & 37.8  & 36.9 & 40.7  & 39.6  & 45.2  & 37.4 & 34.5   & 46.9 & 50.1  & 40.5  & 36.1 & 41.0  & 29.6  & 32.3   & 39.0 \\
Lin \textit{et al.}~\cite{lin2019trajectory} &  {BMVC’19} & 32.5 & 35.3  & 34.3 & 36.2  & 37.8  & 43.0  & 33.0 & 32.2   & 45.7 & 51.8  & 38.4  & 32.8 & 37.5  & 25.8  & 28.9   & 36.8 \\
Pavllo \textit{et al.}~\cite{pavllo20193d} & {CVPR’19}   & 34.1 & 36.1  & 34.4 & 37.2  & 36.4  & 42.2  & 34.4 & 33.6   & 45.0 & 52.5  & 37.4  & 33.8 & 37.8  & 25.6  & 27.3   & 36.5 \\
Liu \textit{et al.}~\cite{liu2020attention} & {CVPR’20}   &{\color{blue}32.3}& 35.2  & 33.3 & 35.8  & 35.9  & 41.5  & 33.2 & 32.7   & 44.6 & 50.9  & 37.0  & 32.4 & 37.0  & 25.2  & 27.2   & 35.6 \\
UGCN~\cite{wang2020motion} & {ECCV’20} & 32.9 & 35.2  & 35.6 &{\color{blue}34.4}& 36.4  & 42.7  &{\color{blue}31.2}& 32.5   & 45.6 & 50.2  & 37.3  & 32.8 & 36.3  & 26.0  &{\color{blue}23.9} & 35.5 \\
Chen \textit{et al.}~\cite{chen2021anatomy} &{TCSVT’21}   & 33.1 & 35.3  & 33.4 & 35.9  & 36.1  & 41.7  & 32.8 & 33.3   &{\color{blue}42.6}& 49.4  & 37.0  & 32.7 & 36.5  & 25.5  & 27.9   & 35.6 \\
\rowcolor{gray!25}
PoseFormer~\cite{zheng2021_3d_hpe} &ICCV’21  & 32.5 &{\color{blue}34.8}&{\color{blue}32.6}& 34.6  &{\color{blue}35.3}&{\color{blue}39.5}& 32.1 &{\color{blue}32.0}& 42.8 & {\color{blue}48.5}&{\color{blue}34.8}&{\color{blue}32.4}&{\color{blue}35.3}&{\color{blue}24.5}& 26.0   &{\color{blue}34.6}\\
\rowcolor{gray!25}
Ours         &         & {\color{red}30.5} & {\color{red}30.5}  &{\color{red}30.5} & {\color{red}30.5}  &{\color{red}30.5}  &{\color{red}30.5}  &{\color{red}28.8} &{\color{red}28.1}  &{\color{red}39.2} &{\color{red}46.3}  &{\color{red}31.6}  &{\color{red}28.2} &{\color{red}31.0}  &{\color{red}20.7}  &{\color{red}22.5}   &{\color{red}30.5}\\
\hline
\label{table2}
\end{tabular}}
\end{table*}

\section{Experiments}

\subsection{Datasets}

\textbf{Human3.6M}~\cite{ionescu2013human3} is a large public dataset for 3D human pose estimation research, featuring $3.6$ million images with corresponding 3D human poses. The dataset includes $11$ professional actors ($6$ male, $5$ female) and spans $7$ scenes (such as discussions, smoking, photography, and phone calls). Comprising videos captured by $4$ calibrated high-resolution cameras at $50Hz$, the dataset's labels are derived from precise 3D joint positions and angles obtained from a high-speed motion capture system. We utilize five subjects from the dataset for training and reserve two subjects for testing purposes.

We evaluate our approach on the Human3.6M~\cite{ionescu2013human3} dataset. To comprehensively assess our method, the experiments are primarily divided into three categories. The first category involves evaluating the 2D detection results as our method aims to improve 3D results through enhanced 2D human pose detection. The second category assesses the application of 2D detection results to 3D detection for evaluating the 3D pose results. The third category evaluates the impact of different numbers of input frames on the detection results since our method models temporal relationships based on video frame sequences.

\subsection{Implementation Details}

Our method is implemented using PyTorch~\cite{paszke2017automatic}.
All experiments were conducted on a single NVIDIA RTX 3090 GPU. We selected five subjects from the Human3.6M dataset for training and evaluated the model on two subjects, utilizing an input resolution of $224{\times}224$. The training process spanned $100$ epochs, employing the AdamW~\cite{kingma2014adam} optimizer, cosine decay learning rate scheduler, and linear warm-up for the initial $20$ epochs. The initial learning rate was set to $0.001$, with a weight decay of $0.05$.
During training, three different lengths of sequence frames were chosen, \textit{i.e.}, $f{=}8,f{=}32,f{=}128$. Our spatial module was constructed using a fine-tuned pre-trained model based on the Swin transformer architecture~\cite{liu2021swin}, while the temporal module underwent fine-tuning using a pre-trained vision transformer~\cite{dosovitskiy2020image}.

\subsection{Evaluation of 2D Human Body Detection Results}

\begin{table}[H]
\scriptsize
\renewcommand{\thetable}{I}
\centering
\caption{Evaluation Results for 2D Human Pose}
\resizebox{\linewidth}{!}{
\setlength{\tabcolsep}{5pt}
\begin{tabular}{lccccc}
\hline
\textbf{Method}      & \textbf{AP$\uparrow$}   & \textbf{AR$\uparrow$}   & \textbf{PCK$\uparrow$}  & \textbf{MSE$\downarrow$} & \textbf{Time (s)(one epoch)$\downarrow$} \\ \hline
ViT-S~\cite{dosovitskiy2020image} & {-}   &{-}    &84.8     & 134.5     & \textbf{654.5}\\ 
ViT-B~\cite{dosovitskiy2020image} &{-}    &{-}    &91.2     &84.0     &1097.6\\ 
HRNet-32~\cite{sun2019deep} &76.5   &79.3    &88.1     &{-} &1544.64\\ 
HRNet-48~\cite{sun2019deep} &77.1    &79.9  &88.9     &{-}  &2210.47\\ 
\rowcolor{gray!25}
Baseline(32frames) &79.5    &86.6    &95.4    &74.7   & 755.3\\ 
\rowcolor{gray!25}
Ours(32frames)  & \textbf{91.4}    & \textbf{94.8}    & \textbf{98.6}  & \textbf{33.6}   & 755.4 \\
\hline

\label{table1}
\end{tabular}}
\end{table}

We conducted a quantitative comparison using ViT-S~\cite{dosovitskiy2020image}, ViT-B~\cite{dosovitskiy2020image}, HRNet-32~\cite{sun2019deep}, and HRNet-48~\cite{sun2019deep}. The model excluding the Image Relations Module is designated as the baseline model, and experiments were performed on this baseline to investigate the impact of the Image Relations Module on performance. Training was carried out on these networks using the Human3.6M dataset~\cite{ionescu2013human3}, and evaluation metrics such as AP (Average Precision), AR (Average Recall), PCK (Percentage of Correct Keypoints), and MSE (Mean Squared Error) were employed for the assessment of 2D human body pose. We computed the time taken for each model to train one epoch on the S1 subject as an indicator of the model's inference speed.

From Table~\ref{table1}, it is evident that our approach achieves the highest accuracy with minimal errors. In comparison to the baseline model, our overall model exhibits a higher accuracy, indicating the effectiveness of the Image Relations Module and its significant enhancement of overall performance. Despite not being the fastest in terms of model inference speed, our method demonstrates exceptional cost-effectiveness. It notably improves prediction accuracy without substantially increasing the inference time. Moreover, our model shows a marginal increase in inference time compared to the Baseline model, suggesting that the Image Relations Module does not introduce a significant computational overhead in terms of the inference duration.

\subsection{Evaluation of 3D Human Pose Detection Results}

The 2D human body pose estimation network proposed in this paper is designed to facilitate 3D human body pose estimation. To demonstrate the effectiveness of our network, we utilize the 2D human body pose estimation results as input for 3D human body pose estimation. PoseFormer~\cite{zheng2021_3d_hpe} is employed as the 3D human body pose detection network, and the network is trained following the experimental settings of PoseFormer~\cite{zheng2021_3d_hpe}.

We use MPJPE (Mean Per Joint Position Error)~\cite{zheng2023deep} as the evaluation metric, which represents the Euclidean distance between predicted and ground truth joint positions. Evaluation is conducted on the test set (S9, S11) for $15$ actions. As shown in Table~\ref{table2}, inferences are made for the $15$ actions of two test subjects (S9, S11), and the last column represents the average error for each action.

As shown in Table~\ref{table2}, we utilized the 2D poses generated by our method as input, employing PoseFormer~\cite{zheng2021_3d_hpe} as the 3D pose network. We conducted a quantitative evaluation of the average error for each joint position using MPJPE and P-MPJPE as evaluation criteria. We drew insights from relevant experimental results in PoseFormer~\cite{zheng2021_3d_hpe}. In the table, red indicates the best performance, and blue indicates the second-best. Inference was performed for $15$ actions on two test subjects (S9, S11), with the last column representing the average error for each action. From Table~\ref{table2}, our method exhibits leadership in both evaluation metrics. Compared to the initial PoseFormer~\cite{zheng2021_3d_hpe}, our method reduces the average error by $9\%$, decreasing from $44.3$ to $40.3$. In terms of the MPJPE metric, our method achieves a leading position in almost all $15$ actions, except for the $Sit$ action. However, in the P-MPJPE~\cite{zheng2023deep} metric, our method achieves the optimal result for all actions, demonstrating an enhancement in 3D human body pose estimation effectiveness.

\begin{figure*}[t]
  \centering
  \includegraphics[width=1\textwidth]{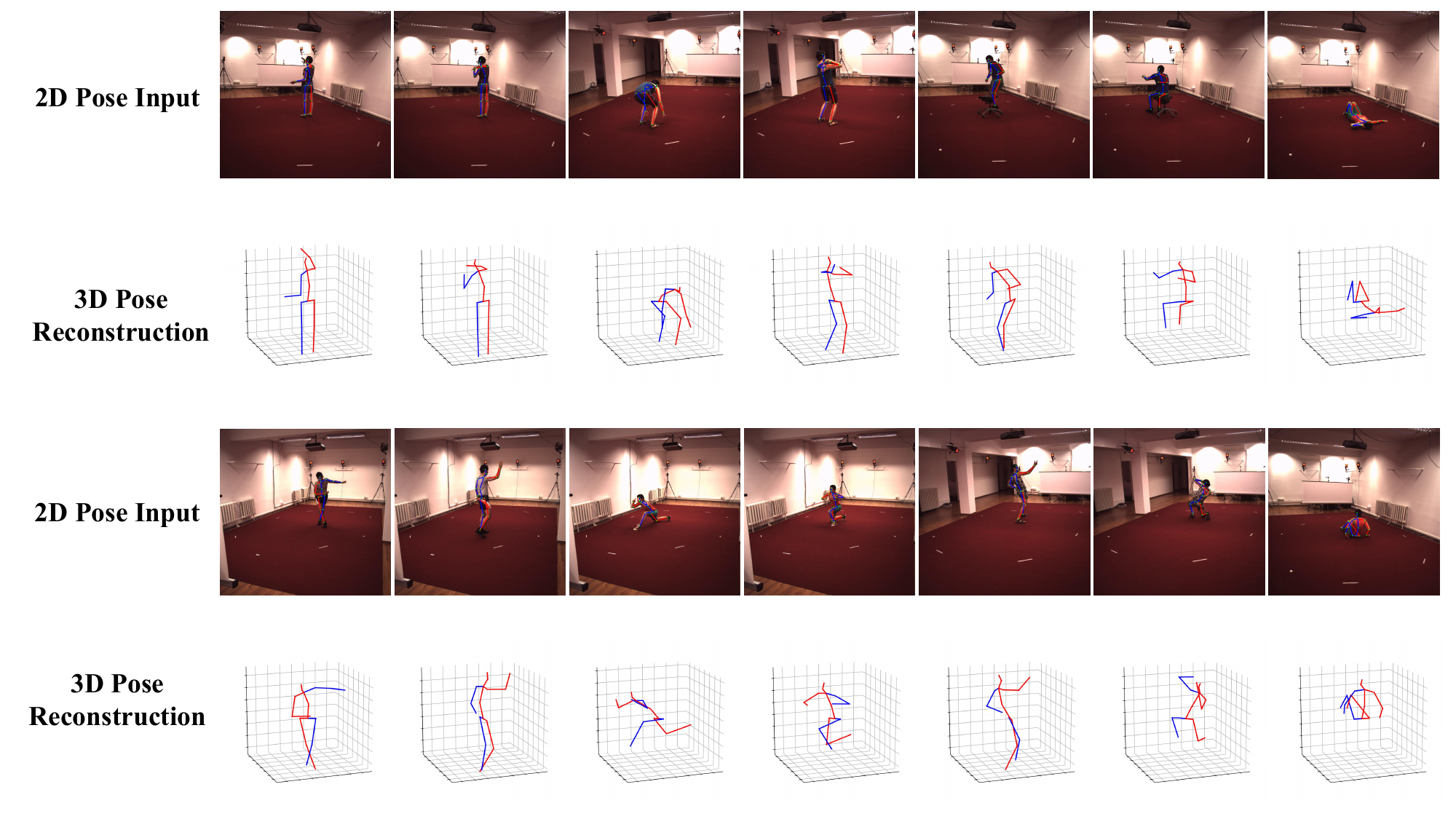} %
  \caption{The detection outcomes of our methodology involve the examination of all images within the Human3.6M dataset. Our approach generates 2D poses for these images, which are subsequently employed as inputs to the 3D Pose detection network, PoseFormer~\cite{zheng2021_3d_hpe}. The network then produces corresponding 3D poses.} %
  \label{img4} %
  \vspace{-0.3cm}
\end{figure*}

As shown in Fig~\ref{img4}, our method detects all images in the Human3.6M dataset and outputs 2D human body poses for all images. Many complex actions are accurately detected, and from Fig~\ref{img4}, we can see that many occluded human keypoints are accurately predicted, thanks to the picture relation module in our network, which can predict these occluded human keypoints based on temporal relations and 3D spatial location relations. We input all the predicted 2D human body poses into the 3D human body detection network Poseformer to reconstruct the 3D human body poses.

\subsection{Frame Sequence Length Analysis}

\begin{table}[H]
\tiny
\centering
\caption{Frame Sequence Length Evaluation}
\resizebox{\linewidth}{!}{
\setlength{\tabcolsep}{8pt}
\begin{tabular}{lcccc}
\hline
\textbf{Frame} & \textbf{AP$\uparrow$}   & \textbf{AR$\uparrow$}   & \textbf{PCK$\uparrow$}  & \textbf{MSE$\downarrow$} \\ \hline
8     & 90.3 & 94.1 & 97.7 & 40.2 \\ 
32    & 91.4 & 94.8 & 98.6 & 33.6 \\ 
\rowcolor{gray!25}
128   & \textbf{92.1}     & \textbf{95.3}      & \textbf{98.9}      & \textbf{28.7}      \\ \hline

\label{table3}
\end{tabular}}
\end{table}

Most 3D pose detection is based on video datasets~\cite{mehta2017monocular,ionescu2013human3,sigal2010humaneva}, so we want the network to be able to process information from an entire video and learn the relationships between all video frames. We set up three different lengths of input frame sequences, $f{=}8, f{=}32, f{=}128$ to evaluate the effect of frame sequence length on the network. In Table~\ref{table3}, we show the effect of the number of frames on the experiment. As the number of frames increases, the accuracy improves, indicating that the longer the input video sequence, the easier it is to capture the relationship between the frame images, resulting in better overall performance. This result proves that our network is able to model the relationships between long sequences of inputs, extract the temporal relationships in them, and achieve better output.

\section{Conclusion}
In this paper, we have looked into multi-perspective temporal-relational 3D human pose estimation and proposed a 2D human body pose detection network that incorporates temporal and spatial relationships to enhance 3D human body pose detection. 
The spatial module is designed to extract features from individual images, whereas an Image Relations Module is established for the global modeling of relationships between images.
The Image Relations Module not only captures temporal relationships between video frames but also learns 3D spatial positional relationships.
Extensive experimental results demonstrate that our proposed approach not only achieves state-of-the-art performance in 2D human body pose estimation but also significantly enhances the effectiveness of 3D human body pose estimation. We also investigated the impact of the length of video frame sequences on our approach and observed an improvement in accuracy as the length of the input video frame sequences increased.

The future work of this study will focus on two main directions. Firstly, we intend to investigate pruning techniques for Transformers to reduce the computational complexity of self-attention, thereby enhancing the modeling of spatial-temporal context in an efficient manner. We aim to process entire video sequences through the network efficiently, meeting the real-time processing requirements. Secondly, we plan to integrate spatial geometry and self-attention mechanisms to better model the 3D spatial relationships within video frames, leading to improved output results for 3D human body pose estimation.

\balance
\bibliographystyle{IEEEtran}
\bibliography{bib}

\end{document}